\documentclass{article}
\usepackage{iclr2027_conference,times}

\usepackage{amsmath,amsfonts,bm}

\def\eqref#1{equation~\ref{#1}}

\def\1{\bm{1}}

\DeclareMathAlphabet{\mathsfit}{\encodingdefault}{\sfdefault}{m}{sl}
\SetMathAlphabet{\mathsfit}{bold}{\encodingdefault}{\sfdefault}{bx}{n}

\usepackage{color,xcolor}
\usepackage{epsfig}
\usepackage{graphicx}

\usepackage{adjustbox}
\usepackage{array}
\usepackage{booktabs}
\usepackage{colortbl}
\usepackage{wrapfig}
\usepackage{hhline}
\usepackage{multirow}

\usepackage{amsmath,amsfonts,amssymb}
\usepackage{bm}
\usepackage{nicefrac}
\usepackage{microtype}
\usepackage{mathtools}

\usepackage{changepage}
\usepackage{extramarks}
\usepackage{fancyhdr}
\usepackage{lastpage}
\usepackage{setspace}
\usepackage{soul}
\usepackage{xspace}

\usepackage[pagebackref=true,breaklinks=true,colorlinks,citecolor=gray]{hyperref}

\usepackage{url}

\usepackage{enumerate}
\usepackage{enumitem}  
\usepackage{makecell}

\usepackage{pifont} 

\usepackage{float}
\usepackage{placeins}
\IfFileExists{algorithm.sty}{\usepackage{algorithm}}{%
  \floatstyle{ruled}%
  \newfloat{algorithm}{tbp}{loa}%
  \floatname{algorithm}{Algorithm}%
  \floatstyle{plain}}
\IfFileExists{algpseudocode.sty}{\usepackage{algpseudocode}%
  }{}
\usepackage{amsthm}
\usepackage[bottom]{footmisc}

\usepackage{listings}
\usepackage{xcolor}
\newcolumntype{L}[1]{>{\raggedright\let\newline\\\arraybackslash\hspace{0pt}}m{#1}}
\newcolumntype{C}[1]{>{\centering\let\newline\\\arraybackslash\hspace{0pt}}m{#1}}
\newcolumntype{R}[1]{>{\raggedleft\let\newline\\\arraybackslash\hspace{0pt}}m{#1}}

\newcommand{\ignore}[1]{}

\DeclareMathAlphabet{\mathbfit}{OML}{cmm}{b}{it}

\makeatletter
\DeclareRobustCommand\onedot{\futurelet\@let@token\@onedot}
\def\@onedot{\ifx\@let@token.\else.\null\fi\xspace}

\def\eg{e.g\onedot} 
\def\ie{i.e\onedot}

\makeatother

\definecolor{MyDarkBlue}{rgb}{0,0.08,1}
\definecolor{MyAqua}{rgb}{0,0.7,0.7}
\definecolor{MyDarkGreen}{rgb}{0.02,0.6,0.02}
\definecolor{MyDarkRed}{rgb}{0.8,0.02,0.02}
\definecolor{MyDarkOrange}{rgb}{0.40,0.2,0.02}
\definecolor{MyPurple}{RGB}{111,0,255}
\definecolor{MyRed}{rgb}{1.0,0.0,0.0}
\definecolor{MyGold}{rgb}{0.75,0.6,0.12}
\definecolor{MyDarkgray}{rgb}{0.66, 0.66, 0.66}
\definecolor{JiayuanColor}{rgb}{0.60,0.43,0.48}

\newcommand{\xhdr}[1]{{\noindent\bfseries #1}}

\newcommand{\squishlist}{
    \begin{list}
    { 
        \setlength{\itemsep}{0pt}
        \setlength{\parsep}{1pt}
        \setlength{\topsep}{1pt}
        \setlength{\partopsep}{0pt}
        \setlength{\leftmargin}{1em}
        \setlength{\labelwidth}{1em}
        \setlength{\labelsep}{0.5em} 
    }
}

\newcommand{\squishend}{
  \end{list}  
}

\definecolor{LightCyan}{rgb}{0.88,1,1}

\usepackage{soul}

\newcommand{\kw}[1]{{\textsc{\MakeLowercase{#1}}}}

\newcommand{\ours}{\kw{Ours}\xspace}
\newcommand{\humandemo}{\kw{Human}\xspace}  
\newcommand{\cop}{\kw{CoP}\xspace}          
\newcommand{\dsrl}{\kw{DSRL}\xspace}        

\IfFileExists{MnSymbol.sty}{\usepackage{MnSymbol}}{}
\usepackage[capitalise]{cleveref}
\usepackage{caption}
\usepackage{subcaption}
\IfFileExists{dsfont.sty}{\usepackage{dsfont}}{}
\usepackage{tikz}

\title{Skill-Space Shooting for Autonomous Robot Policy Improvement}

\author{Zihang Rui$^1$\thanks{Equal contribution. Order determined by coin flip.} \quad Renhao Wang$^2$\footnotemark[1]  \quad Haoxu Huang$^1$ \quad  Yang Gao$^{1,3}$ \\
$^1$Tsinghua University \quad $^2$UC Berkeley \quad
$^3$ Shanghai Qi Zhi Institute }

\iclrfinalcopy 

\begin{document}

\maketitle
\suppressfloats[t]

\begin{abstract}
Robots deployed in the physical world must be able to improve beyond their initial training as they encounter new situations and failures.
For this improvement to scale across tasks, it must make effective use of experience without requiring human demonstration of each correction. Recent agentic systems offer a way to reduce this reliance on human effort by using foundation models to autonomously compose learned behaviors to complete tasks. Yet completing tasks this way does not itself teach a task policy to overcome its own failures; that requires turning these behaviors into learnable corrections for the policy. Our insight is that many such corrections are familiar short behaviors, or skills: they recur across tasks and describe actions that foundation models can reason about from a scene.
We introduce \emph{skill-space shooting}, which uses foundation model guidance to explore corrections through these reusable skills and turn successful trials into policy improvement.
Real-world experiments show repeated improvement in policies acting autonomously, while skills can also be shared to reduce the teaching needed to improve on new tasks.
By making reusable skills a source of corrective supervision, skill-space shooting enables scalable and generalizable policy improvement within and across tasks.
Additional results and videos at \href{https://skill-space-shooting.github.io}{skill-space-shooting.github.io}.
\end{abstract}

\begin{figure}[t]
  \centering
  \includegraphics[width=\textwidth]{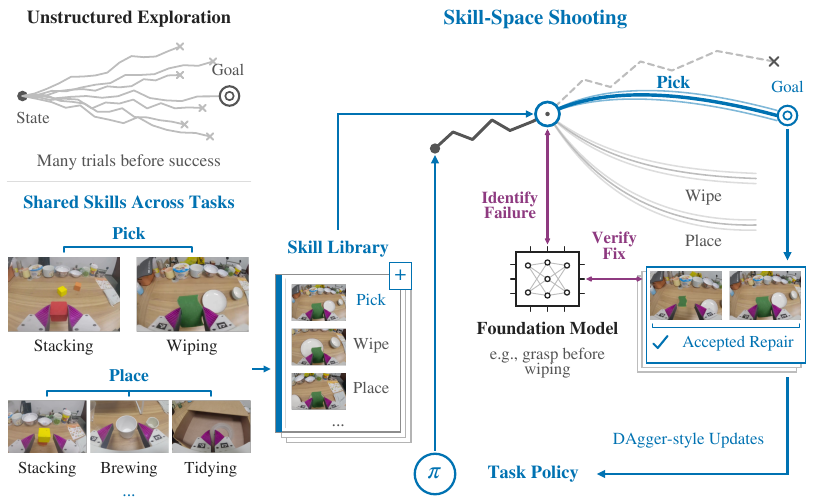}
  \vspace{-1.5em}
  \caption{\textbf{Skill-space shooting uses reusable skills to search for corrections that improve a task
  policy.} We treat learned skills as candidate continuations from a state the policy
  has reached. We employ "shooting" across these skills, with foundation-model knowledge guiding which behavior to try and whether it successfully repairs the failure. Accepted repairs then train the policy for future attempts, closing the improvement loop. Sharing skills across tasks extends this source of corrective supervision while reducing the need for
  new teaching.}
  \vspace{-1em}
  \label{fig:teaser}
\end{figure}

\section{Introduction}
\label{sec:intro}

Scaling robots to a wider range of real-world tasks requires a way for their policies to keep improving after deployment, without relying on continual human guidance.
New objects, changes in the scene, and small execution errors can all expose weaknesses in a task policy.
Simply making more attempts does not provide the experience needed for improvement, as a robot can repeat the same mistake without discovering the right fix.
While humans can supply such fixes, asking for a new demonstration or correction whenever the robot encounters a failure bottlenecks progress to supervision.
We therefore seek an improvement process that generates its own corrective experience and makes initial teaching useful beyond a single task.
Such a process would reduce both the supervision needed to improve a task policy, and the additional human effort needed for generalization to new tasks.

Several established approaches offer appealing routes to this type of improvement loop.
Reinforcement learning uses rewards to improve through interaction and can concentrate exploration around the behaviors of a pretrained policy~\citep{dsrl2025,pld2026}.
This makes further learning possible without asking a person to demonstrate every desired action. But sparse rewards on long-horizon tasks can make the required physical exploration too costly.
Interactive imitation learning (\eg DAgger) takes a more direct route: an expert supplies corrective behavior at states the learner actually visits, and the policy trains on those corrections~\citep{ross2011dagger,liu2023sirius}.
Obtaining these corrections from people, however, imposes a recurring cost as the range of tasks grows.
More recently, agentic robotics has connected foundation-model reasoning to executable skills and programs, allowing knowledge about objects and task structure to guide physical behavior \citep{saycan2022,codeaspolicies2022,programport2023}.
These capabilities make model guidance a plausible resource for discovering better behavior, alongside reward-driven exploration and expert supervision.
However, coordinating skills to complete an assisted execution does not by itself improve the task policy that needed help.
The challenge is thus to use those capabilities to generate corrections at the policy's own failures, so that successful assistance also becomes supervision for learning.

Our key insight is that seemingly disparate long-horizon tasks share many short-horizon behaviors, or skills, that can also supply local corrections.
Separately taught skills can provide useful behavior even when the task policy cannot yet perform the entire task reliably.
For example, a dedicated picking skill only needs to solve a local grasping problem to supply useful actions within a longer robotic kitting task that a policy might not yet complete end-to-end.
Invoking that skill from the policy's current state also addresses the conditions produced by its earlier actions, which a fresh demonstration may not cover.
Foundation models can then reason over which of these skills to try, and monitor physical rollouts to test whether the proposed correction works.
A successful trial therefore supplies both a way to continue the task and actions that teach the policy how to handle its own difficulties.
This turns the skill into a source of autonomous supervision, allowing further trials to address the remaining failures as the policy improves.
Because that skill also need not be unique to one task, sharing its training data can also reduce the teaching needed to prepare corrections for another distinct task.

We develop this approach as \emph{skill-space shooting}: autonomous trials of learned skills as candidate continuations of a task policy's rollout.
Our central claim is that these skills can be invoked autonomously, stitched into a learner's rollouts to provide corrective supervision, and reused through a shared library to support improvement across tasks.
A learned failure detector decides when to invoke a skill, while a foundation model uses semantic knowledge to select the skill and check the repair.
Accepted repair segments from completed rollouts then update the task policy through DAgger-style aggregation.
Across four real-world tasks, skill-space shooting improves unaided policy performance over successive iterations, including from a policy with no observed initial task successes.
Our intervention analysis also shows that the library supplies useful repairs at the policy's failure states.
Finally, we demonstrate that sharing skill data reduces the additional human effort needed to adapt existing skills to new motion primitives, and even facilitates improvement on a new task.
Together, these results support skill-space shooting as a way to make initial human teaching reusable for autonomous policy improvement within and across tasks.

\section{Related Work}
\label{sec:related}

We contextualize skill-space shooting against three main bodies of work, ranging from skill composition, to learning from corrections, and finally to autonomous improvement from experience.

\paragraph{Agentic composition of primitives.}
Agentic robotics connects semantic reasoning to physical behavior by giving models access
to executable skills and programs~\citep{codeaspolicies2022}.
Temporally extended behaviors have long served as actions for learning and planning, as formalized by options in~\citet{sutton1999options}. Recent agentic systems add semantic guidance: SayCan grounds language-based skill selection in learned feasibility, while
CoPa grounds manipulation through spatial constraints~\citep{saycan2022,copa2024}.
Programmatic and hierarchical approaches use such connections between reasoning and action
to execute sequences of behaviors~\citep{codeaspolicies2022,programport2023,hirobot2025}.
Skill libraries can also be mined from prior data to support adaptation to new tasks, as in
SkillPlug~\citep{skillplug2026}. Execution feedback can further improve these systems: ASPIRE revises programs and stores repair knowledge, while RoboClaw refines the skill policies its agent composes~\citep{aspire2026,roboclaw2026}. Online recovery further improves execution around a fixed task policy~\citep{flare2026,criticloop2026,recovla2026,failsafe2025,probeact2026,toerr2024}. Together, these methods make learned behaviors available to model guidance during execution. Our learning problem is to make this assistance improve the complete-task policy itself.
We use skill executions as corrective supervision, so the policy can retain their benefit when the composition and repair system is removed.

\paragraph{Learning from corrections.}
Interactive imitation learning addresses the states an imperfect policy actually reaches,
including situations absent from its original demonstrations. The classic approach is DAgger: one collects expert behavior at these failure states and aggregates it to train successive policies~\citep{ross2011dagger}.
Later work improves how this supervision is weighted and reused~\citep{mandlekar2020iwr,liu2023sirius,rac2025,pistar06recap}, or enables taking corrections in language~\citep{yay2024,rth2024,droc2023}, or learns when such help is needed~\citep{thriftydagger2021,aim2025,valueformer2026}.
These approaches make supervision more useful or easier to supply, motivating further automation of the corrective behavior itself. UniIntervene goes further toward automating intervention during RL, combining a temporal value-risk critic with recovery toward targets retrieved from previous interventions~\citep{uniintervene2026}. Our framework retains the focus on learner-visited states but uses demonstrated primitives to supply corrective actions. Initial teaching thus supplies behavior that the robot can invoke autonomously as the task policy improves, without asking a person to demonstrate each new correction.

\paragraph{Autonomous improvement from experience.}
Autonomously collected experience must somehow provide useful supervision for further improvement~\citep{scaleup2411,belkhale2023datacentric}.
Existing systems use this experience to retrain policies, fit critics and world models,
or learn recovery behaviors~\citep{robocat2023,offpolicyq2608,rise2026,resync2026}. To guide exploration with prior experience, SPiRL learns a state-conditioned skill prior,
while SkiLD combines transferable skills with target-task demonstrations~\citep{spirl2021,skild2022}. DSRL likewise reuses a pretrained policy, optimizing its latent-noise inputs~\citep{dsrl2025}. These approaches already show that exploration can exploit learned structure. Skill-space shooting builds on this to use language-described skills to make candidate corrections accessible to foundation-model reasoning. These skills and accompanying visual and textual descriptions let the model apply its knowledge of objects, actions, and task structure
when choosing which behavior to try. This approach also connects to SOAR, which uses VLMs to collect and assess semantically meaningful experience~\citep{soar2024}. In our work, semantic guidance instead directs physical repair trials at difficulties encountered by the task policy. The resulting behavior must then improve a policy that can act without assistance.

\begin{figure}[t]
  \centering
  \includegraphics[width=\textwidth]{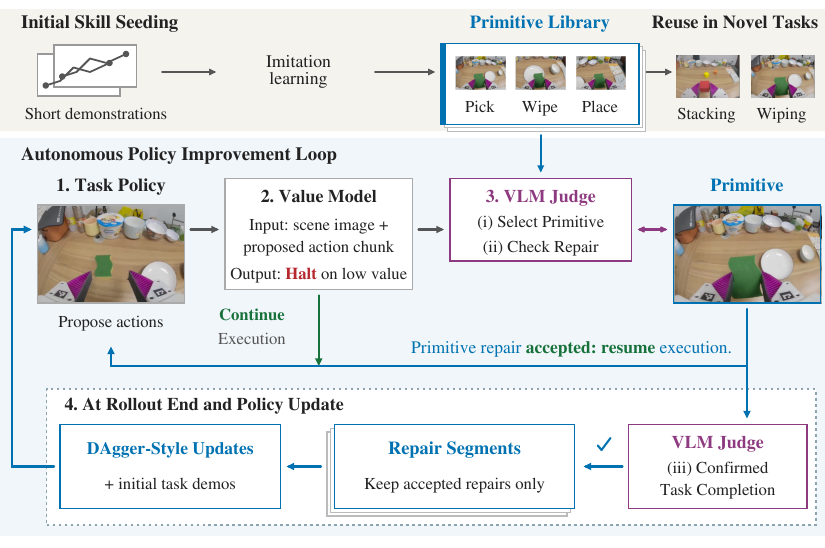}
  \vspace{-1.5em}
  \caption{\textbf{Trials of reusable skills supply corrective supervision for the task policy.}
  A value model detects likely failures, and a VLM selects a skill and verifies its repair before the task policy resumes.
  Accepted repair segments from completed tasks train the next policy, which collects the next round of experience.}
  \vspace{-1em}
  \label{fig:method}
\end{figure}

\section{Method}
\label{sec:method}

\xhdr{Problem setting.}
We begin with an imperfect task policy trained by behavior cloning on initial task demonstrations, together with a library of learned short-horizon skills. For simplicity, we decompose our tasks into stages with a designated skill, or primitive, for each stage. For example, a drawer task might involve three such primitives: opening the drawer, placing an object inside, and closing it. We collect demonstrations of these stages independently to seed the library, allowing each skill to learn its local behavior without having to solve the complete task. Crucially, this initial teaching need not be repeated from scratch for every task: skills and their training data can come from previous tasks and settings, and our own sharing experiments will show how existing skill data reduces the teaching needed to prepare corrections for a new task. We use demonstrations only to establish this starting library; acquiring skills from human video or autonomous experience offers further work directions for extending it. Given the initial task policy and library, our focused contribution is instead how to turn these skills into corrective experience that improves the policy.

Skill-space shooting uses these learned skills to try corrections where a task policy is likely to fail, then trains the policy on the successful repairs. Our key insight is that a skill can supply a useful local behavior even when the policy cannot yet complete the whole task.
To turn that capability into supervision, the loop must recognize when help is needed, test a correction from the learner's current state, and preserve the actions that resolve the difficulty (\cref{fig:method}).
The rest of this section describes how the models identify failure to initiate these shooting trials, how repairs become supervision over successive iterations, and how a shared skill library supports improvement on new tasks.

\subsection{Deciding When and How to Shoot}
\label{sec:invoke}

\xhdr{Detecting a likely failure.}
The task policy controls the robot until it needs a correction, so the first decision is when to interrupt its execution. Waiting for an obvious failure can leave little opportunity to recover: an object may appear securely grasped immediately before a motion drops it.
The detector must therefore consider what the robot is about to do as well as what it sees.
We score the current observation $o_t$ and the policy's proposed $H$-step action chunk $a_{t:t+H-1}$, and trigger a trial when
\begin{equation}
  V_\phi(o_t, a_{t:t+H-1}) < \tau.
  \label{eq:trigger}
\end{equation}
for some calibrated threshold $\tau$. In our implementation, DINOv2 image features~\citep{dinov2} and encoded action tokens are used instead of raw observations and actions, respectively. These are then used to condition a flow-matching head, which we take to be the parameterization of $V_\phi$~\citep{lipman2023flow}.
The head predicts the post-chunk robot state together with a scalar value, which determines the trigger condition. We train this value with targets of $+0.5$ for successful execution and $-0.5$ beginning one action horizon before a recorded failure. These targets encourage the detector to flag an approaching failure while a skill can still correct it. We use $H=16$ and the midpoint threshold $\tau=0$ on every task, removing any need for hyperparameter sweeping. Each task's detector is trained on initial labeled episodes and then frozen. This facilitates the fully autonomous improvement loop (beyond initial human demonstrations and subsequent scene resets).

\xhdr{Selecting the skill correction.}
Once a low value flags a risky motion, we search for a corrective continuation in the space of learned skills. Each skill in the library defines a candidate behavior to try from the state the task policy has reached, such as recovering a grasp, finishing a placement, or repeating a sweep. Searching over these recognizable behaviors allows a foundation model to use its knowledge of objects and task structure to guide which continuation to attempt. We instantiate Gemini 2.5 Pro~\citep{gemini25} as this foundation model judge, providing it with the current scene, task instruction, and available skills. For example, if the sponge is still in its bowl when the policy attempts to wipe, the judge can identify the missing grasp and select the corresponding skill. The selected skill then supplies the coordinated actions for a physical trial, whose outcome determines whether the proposed correction works. In this way, foundation-model reasoning guides \emph{how} to shoot, while the value model determines \emph{when} a trial is needed. 

In our implementation, recall again that for simplicity each task stage has a designated primitive. This means that in practice identifying the stage that needs recovery or correction also determines exactly which skill to shoot. For scalability, the task and skill descriptions are inputs to a fixed prompt, allowing the same selection procedure to apply across tasks.

\subsection{Repairing Rollouts and Improving Policies}
\label{sec:stitch}

\xhdr{Shooting a skill.}
To test a correction behavior, the selected skill takes control where the task policy stopped. Earlier errors may have changed object positions or contacts, so the skill must respond to the state the task policy actually reached. This is also what makes the resulting actions useful supervision: a fresh demonstration from a clean starting state may never encounter the same difficult failure state. Once the stage is repaired, control returns to the task policy so it can attempt the remaining stages. Returning control lets the task policy use the behaviors it already performs well and encounter further difficulties that may need correction within the same rollout.

Because selecting a skill does not guarantee a successful repair, the loop checks its physical outcome before returning control. The value model monitors the skill during shooting, signaling when to request this check. After $c$ consecutive nonnegative checks, or a limit of $M$ executed actions, the foundation model checks whether the stage was repaired. An accepted repair lets the task policy resume; otherwise, the same skill retries from the current state. We allow at most three retries per rollout and discard it if this budget is exhausted without a repair. We use $c=8$ on Stack-3, Drawer, and Sweeping, and $c=16$ on Coffee, whose skills take longer; $M=100$ throughout.

\xhdr{Learning from repairs.}
The repaired rollout contains actions for the task policy's failure states, but also the mistakes that led there. Training on the full sequence would teach the policy to imitate both. We therefore retain only accepted, skill-executed repair segments from rollouts the judge confirms as complete. Each segment contains observations and executed actions from invocation to handback, relabeled with the task instruction. We exclude the policy's actions and rejected trials.

This produces DAgger-style supervision \citep{ross2011dagger} at states visited by the learner, with demonstrated skills supplying the corrective actions.
In particular, training need not wait for the policy's first end-to-end success: a completed rollout can contain stages that only the skills could perform.
At iteration $k$, let $\mathcal{R}_k$ contain the retained repairs and $\mathcal{D}_0$ the initial task demonstrations.
We update the policy from its previous checkpoint by imitation learning on
\begin{equation}
  \mathcal{D}_{k+1} = \mathcal{D}_0 \cup \bigcup_{j=0}^{k}\mathcal{R}_j.
  \label{eq:mix}
\end{equation}
The updated policy then collects the next round of experience. This iteration helps the task policy learn corrections and exposes a wider range of failures. For example, a policy that now completes an early grasp may reach a placement stage that previous attempts never reached. Iteratively collecting with the updated policy brings these newly encountered states, along with remaining failures, to the same skill library. The skills thus provide new supervision as the learner progresses, without needing another round of demonstrations. This process remains autonomous: the models decide when to intervene, which corrections to shoot, and whether to accept them. The process is also scalable and minimizes human involvement: skills and judge prompt remain fixed, as does the value model after its initial calibration, so human input is limited to initial demonstrations and scene resets.                                              

\subsection{Reusing the Skill Library}
\label{sec:share}

\xhdr{Why skills?}
For our improvement loop to work, corrections must be meaningful and identifiable choices for the judge, as well as executable behaviors for the robot. A short-horizon skill satisfies both these conditions: its identity (visual and textual descriptions) describes a clear local correction, while its policy supplies the coordinated actions needed to attempt it. For example, selecting a grasp invokes an approach, gripper closure, and lift, with the skill responding to observations along the way. The foundation model can therefore reason about the behavior needed to repair a stage while the skill handles its physical execution. Teaching each skill separately also lets it learn this shorter behavior without having to master the entire task, which is a far easier learning problem.

In our framework, the skill library includes basic primitives like picking and placing, dynamic and periodic primitives like sweeping, or fine-grained and complex primitives like hanging a broom on a nail. Together, these skills illustrate the range of corrections our interface accommodates. We train these skills from handheld UMI demonstrations~\citep{umi2024}; both the skill and task policies follow from a $\pi_{0.5}$-class VLA architecture~\citep{pi05}. This gives them compatible observation and action interfaces as well as control behaviors, allowing seamless handoffs.

\xhdr{Sharing across tasks.}
Because these behaviors recur across tasks, their training data can help prepare corrections for a new task. For example, grasping a new object may require adapting to its shape while retaining the approach-and-lift behavior learned from earlier objects. To this end, we combine existing skill data with new demonstrations to train the adapted primitive, and show that this reduces the need to teach the entire motion primitive from scratch. The new task has its initial policy and value model, but uses the same procedure to invoke skills, check repairs, and learn from them.

\section{Experiments}
\label{sec:exp}

Our experiments answer three questions about skill-space shooting as a source of reusable corrective supervision.
(1) Do skill-based corrections improve the task policy over successive iterations?
(2) Can the system autonomously invoke skills and verify their repairs reliably?
(3) Can a shared skill library transfer across distinct tasks, reducing the human effort needed to prepare new skills?

\paragraph{Tasks and baselines.}
Our four main tasks require a range of corrective behaviors. \textbf{Stack-3} stacks three cubes one on top of another in sequence, while \textbf{Drawer} requires opening a drawer, placing a roll of tape inside, and closing it. \textbf{Coffee} pours milk, coffee, ice, and adds a stirring stick to a cup. Finally, \textbf{Sweeping} retrieves a small handheld broom, sweeps six blocks into a target zone, and rehangs the broom. Together, these tasks test corrections within ordered sequences and during dynamic contact. 

The baselines test alternatives to collecting corrections at the task policy's failures. \dsrl uses sparse terminal success rewards to learn the latent-noise inputs of the pretrained task policy under a matched collection-time budget for each improvement iteration~\citep{dsrl2025}.
\cop executes the same skills from our library, but under human high level guidance, with up to three rewind-and-retry operations per rollout.
\humandemo adds complete-task demonstrations whose total duration matches our newly retained repair data, using the same policy-update procedure.
These comparisons test exploration, direct skill composition, and further human teaching, respectively.

\paragraph{Evaluation.}
We evaluate the learned policies without the value model, judge, or skill library, so the corrective behaviors must be internalized in the task policy. The only exception is \cop, which is evaluated as a chain of primitives. Our loop collects 21--81 rollouts per round depending on the task, running four rounds on Stack-3, Drawer, and Sweeping, and six on Coffee.
Each method has one run per task, with 30 evaluation rollouts per checkpoint on Stack-3 and 20 on each other task. Stack-3 and Drawer use full-task success; Coffee and Sweeping use mean temporal progress, crediting only consecutive units completed before the first failure.
Coffee has four equal units; Sweeping has eight: broom pickup, six blocks, and broom return. Progress reveals improvements to earlier stages while preserving the dependencies between them. We report bootstrap 95\% confidence intervals for progress and summarize plateau performance by the best of the final three evaluations.
Beyond initial task and skill demonstrations, our loop requires human effort only for scene resets.

\begin{figure}[t]
  \centering
  \includegraphics[width=\textwidth]{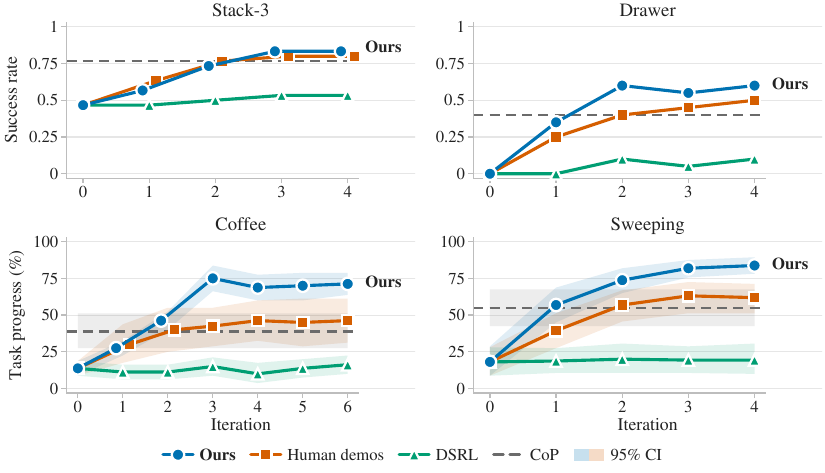}
  \caption{\textbf{Skill-space shooting improves policies over iterations, even from zero observed initial successes.}
  Baselines are latent-space RL (\dsrl), additional demonstrations (\humandemo), and fixed primitive composition (\cop).
  Top panels report full-task success; bottom panels show temporal progress with bootstrap 95\% intervals.}
  \label{fig:main}
\end{figure}

\subsection{Main Results}
\label{sec:exp-improve}

As shown in \cref{fig:main}, successive updates improve unaided performance on all four tasks without new human demonstrations. Because evaluation uses the task policy alone (\ie without guidance from the value model or foundation model), these gains show that the benefit of a skill correction extends beyond the rollout in which it was supplied. The task policy retains behavior from its repairs, allowing subsequent collection rounds to address new difficulties or stubborn error modes. 

Performance on Drawer is particularly interesting. The initial policy succeeds in none of 20 evaluation attempts, making successful experience difficult to obtain through further policy trials. With sparse terminal rewards, exploration must happen upon a sequence that completes the task before receiving a positive task-level signal. Skill-space shooting narrows this search to learned behaviors that already provide competence at individual stages. Foundation-model guidance selects a skill suited to the current difficulty, allowing the robot to try a coherent correction and continue toward later stages. The robot can therefore construct a successful attempt through targeted local repairs, then learn from the actions that made those repairs work. After one update, unaided success reaches 7/20; \dsrl remains at zero after its first update and reaches at most 2/20. This result supports the central motivation for skill-space shooting: existing competence in short behaviors can make successful corrective experience accessible even when exploration through the complete-task policy rarely succeeds.

\begin{table}[!tbp]
  \centering
  \small
  \setlength{\tabcolsep}{6pt}
  \begin{tabular}{@{}lcccc@{}}
    \toprule
    Method
      & \begin{tabular}[c]{@{}c@{}}Stack-3\\{\scriptsize $(N = 30)$}\end{tabular}
      & \begin{tabular}[c]{@{}c@{}}Drawer\\{\scriptsize $(N = 20)$}\end{tabular}
      & \begin{tabular}[c]{@{}c@{}}Coffee (\%)\\{\scriptsize $(N = 20)$}\end{tabular}
      & \begin{tabular}[c]{@{}c@{}}Sweeping (\%)\\{\scriptsize $(N = 20)$}\end{tabular} \\
    \midrule
    \dsrl      & 0.533 & 0.100 & $16.25 \pm 6.25$ & $20.00 \pm 10.63$ \\
    \cop       & 0.767 & 0.400 & $38.75 \pm 12.50$ & $55.00 \pm 12.50$ \\
    \humandemo & 0.800 & 0.500 & $46.25 \pm 15.00$ & $63.13 \pm 11.88$ \\
    \midrule
    \ours      & \textbf{0.833} & \textbf{0.600} & $\mathbf{71.25} \pm 7.50$ & $\mathbf{83.75} \pm 5.63$ \\
    \bottomrule
  \end{tabular}
  \caption{\textbf{Skill-space shooting reaches the highest observed plateau on every task.}
  Best of the last three evaluations; \cop is evaluated once.
  Success rates or mean progress $\pm$ a bound covering its 95\% bootstrap CI.}
  \label{tab:main}
  \vspace{-2mm}
\end{table}

\xhdr{Comparison to baselines.}
Skill-space shooting reaches the highest observed plateau on all four tasks, with its clearest advantages on Coffee and Sweeping (\cref{tab:main}). These tasks demand different kinds of competence: Coffee combines pouring and placement across an extended sequence, whereas Sweeping requires sustained control during contact. The larger gains over additional end-to-end human demonstrations suggest that \emph{where} the new experience is collected matters, particularly when much of a rollout already works in subsequent iterations. The comparisons with \dsrl and \cop further clarify the role of the skills. Our advantage over \dsrl extends beyond Drawer, reaching 55 progress points on Coffee and 63.8 on Sweeping. While \dsrl clearly struggles to effectively explore and learn under sparse rewards, shooting in a structured skill space provides coordinated corrective actions directly, reducing the search needed to discover them through terminal rewards. Yet note that simply executing those skills does not achieve the same final performance: our initial policies begin below \cop, then exceed it on every task while acting without assistance. The library is therefore useful beyond its performance as a standalone skill chain. By stitching its corrections into the task policy's rollouts and training on those segments, the loop lets the task policy retain its existing competence while acquiring behavior for the states where it struggles. Thus, the resulting policy can outperform both its initial behavior and direct composition of the skills that helped improve it.

\begin{table}[t]
  \centering
  \small
  \setlength{\tabcolsep}{3pt}
  \begin{tabular}{@{}lcccc@{}}
    \toprule
    Metric & Stack-3 & Drawer & Coffee & Sweeping \\
    \midrule
    Value model trigger accuracy & 0.75 & 0.75 & 0.83 & 0.71 \\
    Judge stage/skill accuracy & 1.00 & 0.95 & 0.90 & 0.90 \\
    Skill repair success & 0.70 & 0.70 & 0.77 & 0.74 \\
    Judge completion accuracy & 0.95 & 0.95 & 0.90 & 0.95 \\
    \bottomrule
  \end{tabular}
  \caption{\textbf{Reliable interventions and repairs on all four tasks.}
  All use one trigger threshold and judge prompt.}
  \label{tab:invocation}
  \vspace{-1em}
\end{table}

\subsection{Reliability of Interventions and Repairs}
\label{sec:exp-audit}

\xhdr{Can the models identify and verify useful corrections?}
For autonomous trials to supply supervision, the robot must recognize where help is needed and determine whether the attempted correction worked. \Cref{tab:invocation} analyzes these decision junctures.
The first two rows support using model guidance to initiate targeted skill trials. The value model achieves 71--83\% trigger accuracy, while the judge identifies the appropriate stage or primitive with 90--100\% accuracy. Together, these results indicate that the system can usually recognize a need for correction and connect it to a relevant skill in the library, without a person deciding when or how to intervene. Skills also successfully repair 70--77\% of correctly triggered failure cases. Note that our retry budget gives a plausible skill multiple opportunities to succeed. Inside of three independently but identically distributed attempts, we should be able to repair failures successfully with probability 97.3--98.8\%. Finally, the end of the improvement loop still needs to verify the outcome of each trial and checks overall task completion before retaining repair segments. With completion-judgment accuracy of 90--95\%, this final check is remarkably robust.

\begin{figure}[!tbp]
  \centering
  \includegraphics[width=\textwidth]{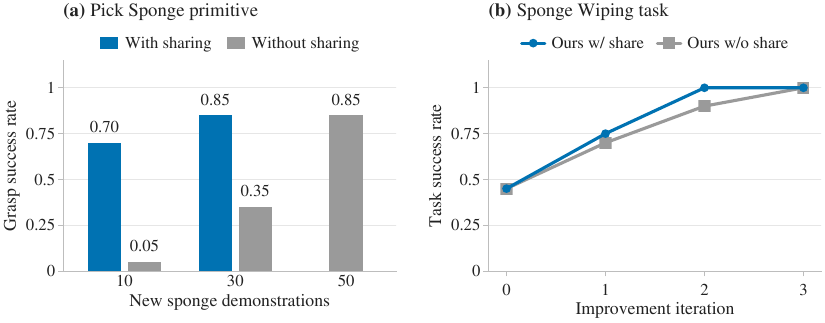}
  \vspace{-1.5em}
  \caption{\textbf{Sharing skill data reduces new teaching and supports policy improvement on new tasks.}
  Left: cube-picking data improves sponge grasping at matched new-demonstration counts.
  Right: shared pick-and-place primitives support Sponge Wiping improvement with fewer new demonstrations than unshared primitives.}
  \vspace{-1em}
  \label{fig:sharing}
\end{figure}

\subsection{Sharing and Extending Skills For New Tasks}
\label{sec:exp-share}

\xhdr{Can we adapt existing skills to new affordances?}
Scaling the loop across tasks requires making existing skill data useful beyond their original objects.
We test this with \textbf{Pick Sponge}, combining cube-picking demonstrations from Stack-3 with new demonstrations of grasping a deformable sponge (\cref{fig:sharing}a). In the shared condition, we finetune a checkpoint pretrained on 150 cube-picking demonstrations: 50 for each of three cubes from Stack-3. Without sharing, we finetune directly from the $\pi_{0.5}$-base model. The horizontal axis counts new sponge-picking demonstrations, so each comparison measures the benefit of prior cube experience at the same budget for new demonstrations.

With only 10 new demonstrations, sharing raises grasp success from 5\% to 70\%. With 30, sharing matches the success obtained from nearly double the number of demonstrations without sharing. Prior skill learning therefore does more than improve performance at a fixed data budget: it reduces the new demonstrations needed to prepare the same skill for an unfamiliar object/motion primitive.

\xhdr{Can the transfer skills from one task to support improvement of a different task?}
Next, we establish that an adapted skill can then correct another policy's failures and return control successfully on an entirely new task. Our task of choice is \textbf{Sponge Wiping}: the robot picks up a sponge, wipes a white bowl, and finally places the sponge in a storage receptacle. This task requires three primitives: pick, wipe, and place. With sharing, all three are finetuned from the cube-picking checkpoint, using 10 new pick demonstrations, 50 wipe demonstrations, and 10 place demonstrations. Without sharing, each primitive is finetuned from $\pi_{0.5}$-base using 50 end-to-end demonstrations. Both conditions therefore receive the same amount of wiping data, with sharing reduces the new teaching for picking and placing. We then use these libraries in otherwise identical repair-and-update loops (\cref{fig:sharing}b). Despite requiring fewer new demonstrations to prepare its primitives, the shared condition supports improvement from 45\% to full observed task success, reaching that level one iteration earlier than the unshared condition. 
We thus show that prior skill data can also transfer usefully to improving an entirely new task, further motivating the role of skill libraries.
\section{Limitations and Future Directions}
\label{sec:limits}

\xhdr{Acquiring and expanding skills.} Repeated improvement with a fixed library shows that skills can remain useful after the policy learns from earlier corrections, but specifying and teaching them still requires initial human effort. Foundation models will increasingly help decompose tasks and identify reusable behaviors~\citep{hansen2026geminiroboticser2}. Persistent failures could guide which skills to acquire next, with each addition evaluated by how many tasks it benefits. Learning skills from human video~\citep{mimicplay2023} could further reduce demonstration effort as the library expands.

\xhdr{Sustained autonomous operation.} Although the improvement loop is autonomous, people still reset scenes between rollouts. Learned reset behaviors~\citep{roboclaw2026} or task loops, where completing one task prepares the next~\citep{gupta2021resetfree}, could enable unattended operation over days or weeks on open-ended tasks. Combining them with skill-space shooting would allow sustained collection and test whether corrective supervision remains useful as tasks and surroundings change.

\xhdr{Generalizing failure detection.} Our failure detector requires per-task training, though skill selection and repair verification share a common procedure. Foundation models already detect execution failures, track task progress, and locate events in robot video~\citep{hansen2026geminiroboticser2}. As their temporal reasoning improves, extending these capabilities to anticipate failure from proposed motions could remove the need for a separate detector. A shared model would then decide both when to shoot and which skill to try, reducing preparation for new tasks.
\section{Conclusion}
\label{sec:conclusion}

We introduce skill-space shooting to make reusable short-horizon robot behaviors a renewable source of corrective supervision. Foundation-model guidance identifies meaningful behaviors to shoot at a task policy's failures, and monitors their physical rollout to test whether they provide a correction. Training on accepted repair segments from completed rollouts then incorporates those corrections into the task policy.
Our experiments show repeated gains that persist without skill assistance, including improvement from a policy with zero observed initial successes. We also prove that sharing skill data reduces the new teaching needed to prepare another task's improvement loop, connecting improvement within a task to reuse across tasks. Skill-space shooting thus provides a foundation for autonomous improvement at scale: reusable skills let robots learn from their own failures and extend that learning across tasks, without requiring humans to teach every possible correction.

\bibliography{refs}
\bibliographystyle{iclr2027_conference}

\end{document}